\documentclass[letterpaper, 10 pt, conference]{ieeeconf}  

\IEEEoverridecommandlockouts                              

\usepackage{amsmath} 
\usepackage{amssymb}  
\usepackage{graphicx}
\usepackage{balance}
\usepackage{lipsum}
\usepackage[ruled, linesnumbered]{algorithm2e}
\usepackage{multirow}
\usepackage{xcolor}
\usepackage{booktabs}

\DeclareMathOperator*{\argmin}{arg\,min}
\DeclareMathOperator*{\argmax}{arg\,max}

\definecolor{fei}{rgb}{0.7764705882352941, 0.23529411764705882, 0.3176470588235294}

\SetKwComment{Comment}{// }{}

\title{\LARGE \bf
Autonomous Image-to-Grasp Robotic Suturing Using Reliability-Driven Suture Thread Reconstruction
}

\author{Neelay Joglekar$^{1}$, Fei Liu$^{1}$, Florian Richter$^{1}$, Michael C. Yip$^{1}$
\thanks{$^{1}$ Department of Electrical and Computer Engineering, University of California San Diego, La Jolla, CA 92093, USA {\tt\small \{njogleka, f4liu, frichter, yip\}@ucsd.edu}}%
}

\begin{document}

\maketitle
\thispagestyle{empty}
\pagestyle{empty}

\begin{abstract}
    Automating suturing during robotically-assisted surgery reduces the burden on the operating surgeon, enabling them to focus on making higher-level decisions rather than fatiguing themselves in the numerous intricacies of a surgical procedure. Accurate suture thread reconstruction and grasping are vital prerequisites for suturing, particularly for avoiding entanglement with surgical tools and performing complex thread manipulation. However, such methods must be robust to heavy perceptual degradation resulting from heavy noise and thread feature sparsity from endoscopic images. We develop a reconstruction algorithm that utilizes quadratic programming optimization to fit smooth splines to thread observations, satisfying reliability bounds estimated from measured observation noise. Additionally, we craft a grasping policy that generates gripper trajectories that maximize the probability of a successful grasp. Our full image-to-grasp pipeline is rigorously evaluated with over 400 grasping trials, exhibiting state-of-the-art accuracy. We show that this strategy can be applied to the various techniques in autonomous suture needle manipulation to achieve autonomous surgery in a generalizable way.
\end{abstract}

\section{INTRODUCTION}

Robotic surgical systems grant surgeons fine-grained, tremor-free control of surgical tools, enabling numerous Robotically-Assisted Minimally Invasive Surgery (RAMIS) procedures that promote higher patient outcomes. Although current RAMIS procedures are fully-teleoperated, future surgeries will benefit from autonomous solutions for surgical subtasks, such as tissue manipulation \cite{lin2023superpm, shinde2024jiggle} and blood suction \cite{richter2021autonomous, miao2024hemoset}, to reduce surgeons' fatigue and enable them to focus on higher-level decision-making. Suturing is a particularly tedious subtask that would greatly benefit from an autonomous solution \cite{hubens2003performance, ostrander2024current}. Many prior works over the past two decades have demonstrated needle throwing to various success in structured experimental setups \cite{chiu2021bimanual, chiu2023real, hari2024stitch, kim2024surgical}, leading to an interest in realizing autonomous suturing in more realistic scenarios.
    
A key prerequisite for autonomous suturing in realistic scenarios is suture thread reconstruction and grasping. Thread reconstruction estimates the 3D centerline of suture thread visible in the surgical workspace. Most prior works focus only on suture needle tracking \cite{chiu2023real, jiang2023markerless} and manipulation \cite{chiu2021bimanual, wilcox2022learning, lu2021dual}, but successful suture throws must avoid thread entanglement with the tool or needle \cite{hari2024stitch}; thus, thread reconstruction is crucial for planning such entanglement-free tool paths. Furthermore, grasping suture thread is a very useful manipulation capability, such as in knot tying \cite{chow2014novel} where multiple grasps throughout the length of the thread may be required. 

However, suture thread reconstruction and grasping are difficult to perform due to challenging perceptual conditions. Noisy, light-sensitive endoscopic images effect degradation in depth estimates, and the characteristic pixel sparsity of incredibly-thin suture thread further exacerbates this problem. Plus, there are no definable visual landmarks on the suture thread to saliently collocate keypoints between images. This causes challenges in the application of state-of-the-art 3D reconstruction, such as general stereo matching and deep learning methods \cite{geiger2010efficient, chang2018pyramid}, thereby resulting reconstruction and grasping failures produce insufficient sutures that hinder procedures or even cause further tissue damage to the patient. Hence, it's crucial to explicitly estimate and robustly account for such perceptual degradation.

\begin{figure}[t]
    \centering
    \includegraphics[width=0.95\linewidth]{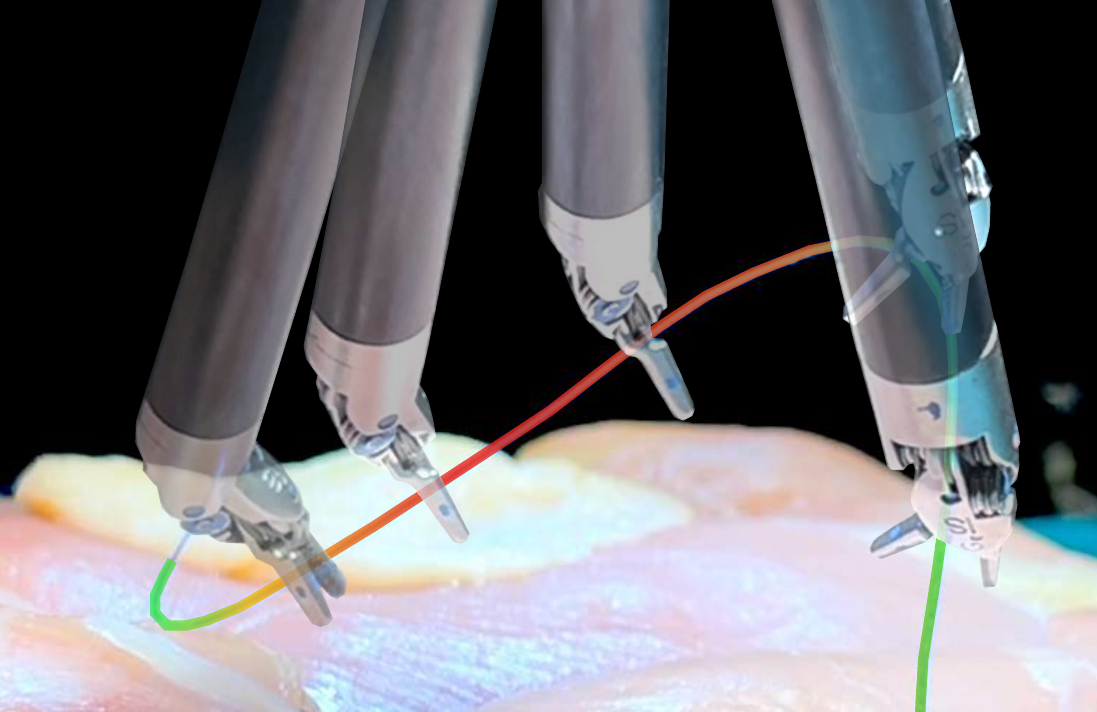}
    \caption{Suture thread reconstruction and grasping are key prerequisites for autonomous suturing, but are difficult to perform due to heavy noise and thread feature sparsity in surgical endoscopic images. We develop a reliability-driven suture thread image-to-grasp pipeline that generates spline reconstructions with attached confidence (where low-to-high confidence is highlighted by the red-to-green spectrum displayed) and executes gripper trajectories that maximize the probability of a successful grasp.}
    \label{title}
    \vspace{-5mm}
\end{figure}

\subsection{Related Works}
``Deformable Linear Object" (DLO) reconstruction is well studied in the computer vision and robotics communities, but applications in surgical scenarios are limited. Most works have been developed for thick ropes and cables and utilize RGB-D cameras that are less noisy than surgical endoscopes \cite{schulman2013tracking, yang2022particle, liu2023robotic}. Jackson et al. and Schorp et al. both developed suture thread reconstruction methods that use stereo matching augmented with pixelwise ordering to construct a 3D pointcloud, to which a NURBS spline is fit \cite{jackson2017real, schorp2023self}. However, neither work explicitly accounts for false pixel matches that result from perceptual degradation and misleading orderings. Lu et al. mitigates this with a ``sampling-based sliding pairing" (SSP) algorithm, which injects extra nodes in a 3D graph that enable their Djikstra's-based reconstruction method to recover from false matches \cite{lu2021toward}. In our previous work, we crafted a keypoint-detection technique that more thoroughly removes noisy matches, along with a spline-fitting optimization that produces smooth, realistic spline reconstructions that satisfy keypoint reliability bounds \cite{joglekar2023suture}. Despite its accuracy and robustness, our spline-fitting optimization required minimizing a complex integral objective, making it inefficient to solve.

Research in suture thread grasping is similarly scarce. Lu et al. additionally provides an algorithm that remedies reconstruction inaccuracy by attempting numerous grasps in a 2D grid of points around a goal until a successful grasp is performed \cite{lu2021toward}. Despite its high success rate, this brute-force approach often requires incredibly long execution times. In our previous work, we proposed utilizing our generated reliability bounds to inform suture thread grasping policies, but we didn't provide such an implementation \cite{joglekar2023suture}.

\subsection{Contributions}

In this work, we achieve state-of-the-art accuracy for suture thread reconstruction and grasping with a reliability-driven approach. In particular, we present the following novel contributions:
\begin{itemize}
    \item A full image-to-grasp pipeline that leverages reliability information gleaned from measured observation noise to robustify suture manipulation.
    \item A new optimization-based reconstruction algorithm that iteratively solves quadratic program (QP) subproblems to more efficiently produce a smooth spline that satisfies reliability bounds.
    \item A grasping policy that utilizes reconstruction confidence estimates to generate gripper trajectories that maximize the probability of grasp success.
\end{itemize}

We evaluate our complete work in a lab setting by performing over 400 grasping trials with the dVRK surgical robotic system on various suture thread configurations and realistic backgrounds. The results from our experiments demonstrate our reconstruction accuracy and grasping robustness.

\section{METHODS}

\begin{figure*}[t]
    \centering
    \vspace{2mm}
    \includegraphics[width=0.98\linewidth]{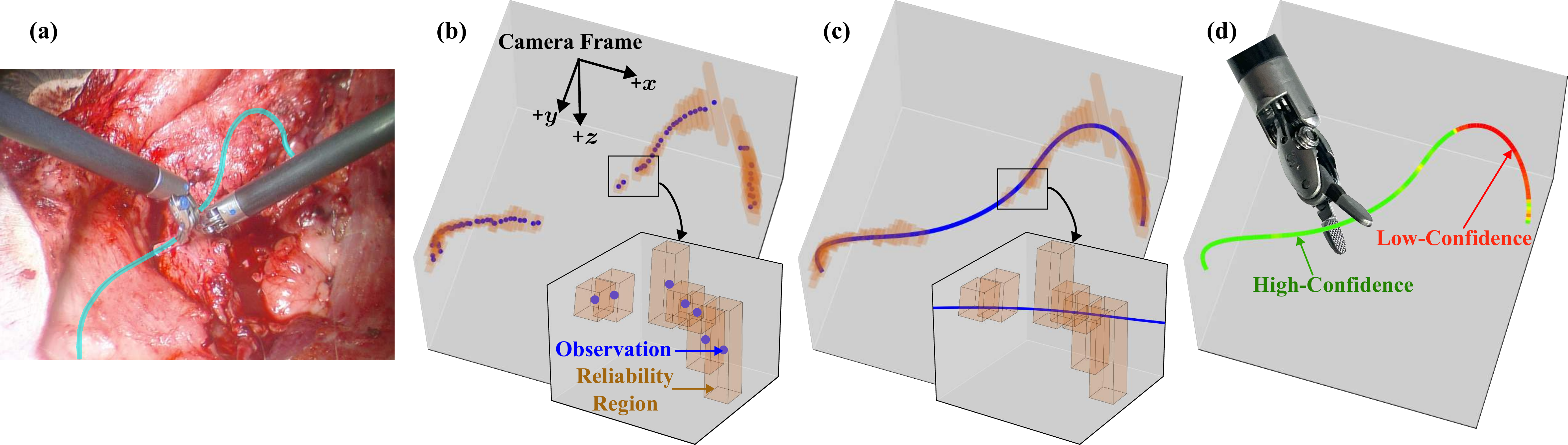}
    \caption{Our full image-to-grasp pipeline: a) Segment the suture thread (cyan) from rectified stereo images. b) Use outlier rejection and clustering to collect thread observations, each with an associated reliability region encoding uncertainty. c) Execute a quadratic-programming-based minimum variation spline reconstruction algorithm to produce a smooth, realistic spline passing through all reliability regions. d) Utilize reliability regions to estimate reconstruction confidence and plan gripper trajectories, maximizing the chance of a successful grasp.}
    \label{pipeline}
    \vspace{-5mm}
\end{figure*}

\begin{figure}[t]
    \centering
    \includegraphics[width=0.95\linewidth]{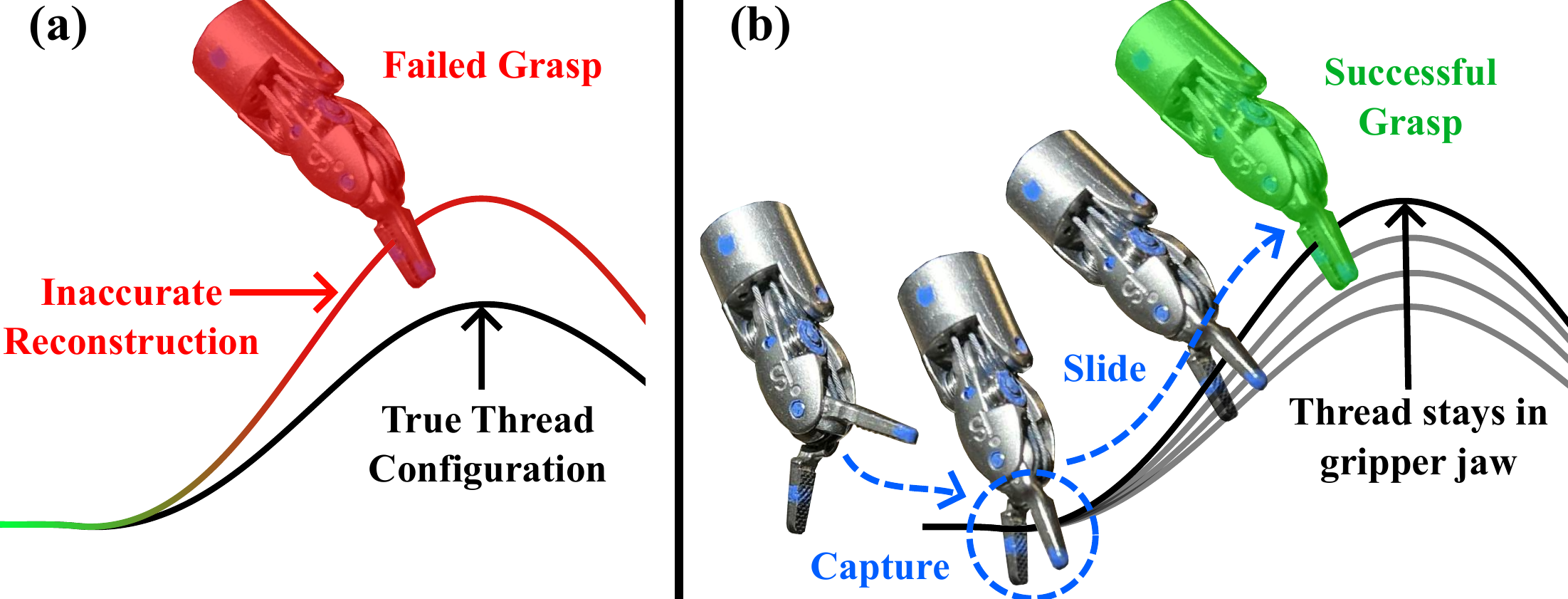}
    \caption{(a) Due to observation noise, some portions of a thread reconstruction may be inaccurate, preventing direct grasping. (b) To mitigate this, we can use estimated reconstruction confidence to initially ``capture" the thread at a reliable point and ``slide" the gripper along our reconstruction to our previously inaccurate grasping point, lightly manipulating the thread such that it stays within the gripper jaws.}
    \label{grasp}
    \vspace{-5mm}
\end{figure}

This section details the implementation of our image-to-grasp pipeline, as is visualized in Fig. \ref{pipeline} and Fig. \ref{grasp}. Additionally, an outline of our reconstruction algorithm is shown in Algorithm \ref{alg:mvs}.

\subsection{Problem Formulation}

We formulate a suture thread curve reconstruction problem that estimates the 3D shape of the thread from stereo endoscopic images. We can initially use the images to measure the positions of discrete points along the suture thread, denoted as the observations $o_1, \dots, o_n \in \mathbb{R}^3$. Due to perceptual degradation, these observations are typically too noisy to be directly used as a thread reconstruction. To account for this noise, each observation $o_j$ is assigned a local reliability region in the camera frame, $\mathbf{R}(o_j) \subset \mathbb{R}^3$. We then want to produce a 3D curve $B(s) : \mathbb{R} \to \mathbb{R}^3$, parameterized by some $s$ defined in the interval $[0, 1]$, that passes through every $\mathbf{R}(o_j)$. $B(s)$ can be represented as a Bezier curve, B-spline, or some other curve object. Since a set of observations may have multiple solution curves, we utilize a smoothing loss $\mathcal{L}(B(s))$ as a regularization technique. The resulting formulation is as follows:
\begin{equation} \label{E:reconstr_formulation}
\begin{aligned}[b]
    &\argmin_{B(s)} \mathcal{L}(B(s))\\
    \textrm{s.t. } &\forall o_j ~\exists s_j : B(s_j) \in \mathbf{R}(o_j)
\end{aligned}
\end{equation}

Additionally, we formulate a grasping policy $\pi (B(s), \mathbf{R}(o_j))$ that utilizes our reconstruction and reliability regions to estimate reconstruction confidence and plan a gripper trajectory that maximizes the probability of a successful grasp.

\begin{figure}[t]
    \vspace{-3mm}
\end{figure}

\begin{algorithm}[t]
    \caption{Iterative QP-based Spline Reconstruction}
    \label{alg:mvs}
    \KwIn{
        Image pair $I_l, I_r$,
        Preset spline parameters $m, d, \{t_k\}_{k=1}^{m+d+1}$
    }
    \KwOut{
        Spline reconstruction $B(s)$, Reliability region parameters $\{s_j\}_{j=1}^{n}$
    }
    
    \Comment{Get observations}
    
    $\{o_j\}_{j=1}^{n} \gets observe(I_l, I_r)$
    \label{alg:getobs}

    \Comment{Compute reliability regions}
    
    $\{\mathbf{R}(o_j)\}_{j=1}^{n} \gets reliabilityRegions(\{o_j\}_{j=1}^{n})$
    \label{alg:getrel}

    \Comment{Initialize $s_j$ values}
    $s_1^{(1)} \gets 0$
    \label{alg:initparams_start}\\
    \For{$j \gets 2 \dots n$}{
        $s_j^{(1)} \gets s_{j-1}^{(1)} + \| o_j - o_{j-1}\|$\\
    }
    $\{s_j^{(1)}\}_{j=1}^{n} \gets \bigg\{\frac{s_j^{(1)}}{s_n^{(1)}}\bigg\}_{j=1}^{n}$
    \label{alg:initparams_end}

    \Comment{Compute QP matrix}
    \label{alg:qpmatrix}
    $\mathbf{A} \gets QPMatrix(m, d, \{t_k\}_{k=1}^{m+d+1})$

    \Comment{Find Minimum Variation Spline}
    $i \gets 1$
    \label{alg:constructmvs_start}\\
    \Repeat{$\left(B^{(i)}\right)'(s) \approx c$}{
        \Comment{Solve QP}
        $\{C_j^{(i)}\}_{j=1}^{n}, \{f_j^{(i)}\}_{j=1}^{n} \gets constraints(\{s_j^{(i)}\}_{j=1}^{n}, \{\mathbf{R}(o_j)\}_{j=1}^{n})$ \label{alg:qpstep_start}\\
        $\mathbf{P}^{(i)} \gets solveQP(\mathbf{A}, \{C_j^{(i)}\}_{j=1}^{n}, \{f_j^{(i)}\}_{j=1}^{n})$\\
        $B^{(i)}(s) \gets spline(\mathbf{P}^{(i)}, m, d, \{t_k\}_{k=1}^{m+d+1})$
        \label{alg:qpstep_end}

        \Comment{Update parameters}
        $\{s_j^{(i+1)}\}_{j=1}^{n} \gets update(\{s_j^{(i)}\}_{j=1}^{n}, B^{(i)}(s))$ \label{alg:updatestep}\\
        $i \gets i+1$
    }
    \label{alg:constructmvs_end}
\end{algorithm}

\subsection{Observations from Endoscopic Images}

We initially segment the endoscopic images with a deep neural network and stereo-match the segmented pixels to form a set of 3D points in the camera frame. For each point $p$ we use a peak-finding algorithm to find the two best stereo matching candidate costs for the associated pixel, denoted as $E_1(p)$ and $E_2(p)$ respectively. These are used for point outlier rejection, based on the following ``peak-similarity" decision metric:
\begin{equation} \label{E:outlier_rejection}
    \sigma\left(\epsilon_1 \left(\frac{E_2(p) - E_1(p)}{\epsilon_2 E_1(p) - \epsilon_3}\right) \right) > \epsilon_4
\end{equation}
where $\sigma$ is the sigmoid function and $\epsilon_1, \epsilon_2, \epsilon_3$ and $\epsilon_4$ are tuning values. The inlier points are clustered and ordered in the pixel space. The associated cluster centroids in the 3D camera frame become our observations $o_1, \dots, o_n \in \mathbb{R}^3$, displayed in Fig. \ref{pipeline}(b). This procedure is implemented in line \ref{alg:getobs} of Algorithm \ref{alg:mvs}.

\subsection{Generating Reliability Regions}

We define each reliability region with a set of bounds centered at each observation. Given accurate image segmentation, we require the left image projection of $\mathbf{R}(o_j)$ to closely match the local segmented region around the projection of $o_j$. We define $\epsilon_{j,u}$ and $\epsilon_{j,v}$ as bounds in the $u$ and $v$ axes of the left camera image centered on the projection of $o_j$, tuned such that they approximate the size of the local segmented region. The size of $\mathbf{R}(o_j)$ should vary based on the reliability of the observed depth of $o_j$. We define a depth bound $\epsilon_{j,z}$ centered on $o_{j,z}$, the $z$-component of $o_j$, based on local variations in depth with other nearby observations:
\begin{equation} \label{E:depth_rel_bounds}
    \epsilon_{j,z} = 1.5 \| L_j(o_{j, z}) - o_{j,z}\|
\end{equation}
where $L_j$ is a least-squares line fitting of the depths of local observations near $o_j$. Note that $\epsilon_{j,z}$ tends to be wider in regions with noisy observations, as shown in Fig. \ref{pipeline} (b). We can utilize these image and depth bounds to define $\mathbf{R}(o_j)$ with the following constraints:
\begin{equation} \label{E:rel_region}
\begin{aligned}[b]
    B(s_j) &\in \mathbf{R}(o_j) \iff\\
    &\begin{vmatrix}
        \frac{1}{B_z(s_j)} K B(s_j) - \frac{1}{o_{j,z}} K o_j
    \end{vmatrix} \leq \begin{bmatrix}
        \epsilon_{j,u} \\
        \epsilon_{j,v}
    \end{bmatrix}\\
    & \begin{vmatrix}
        B_z(s_j) - o_{j,z}
    \end{vmatrix} \leq \epsilon_{j,z}
\end{aligned}
\end{equation}
where $B_z(s_j)$ is the $z$-component of $B(s_j)$ and $K \in \mathbb{R}^{2 \times 3}$ is the left camera projection matrix. These regions are generated in line \ref{alg:getrel} of Algorithm \ref{alg:mvs}.

We split the absolute value constraint into an upper and lower bound, which results in four inequality constraints that are \textit{linear} w.r.t. $B(s_j)$. Note that the image constraints, which use the camera pin-hole projection model, can be converted into a linear constraint by multiplying both sides by $B_z(s_j)$, which is always positive, and factoring out $B(s_j)$ resulting in the following for the upper bound:
\begin{equation} \label{E:rel_linear_ex}
    \left( K - \left(\frac{1}{o_{j,z}} K o_j + \begin{bmatrix}
        \epsilon_{j,u} \\
        \epsilon_{j,v}
    \end{bmatrix} \right)e_3^\top \right) B(s_j) \leq \boldsymbol{0}
\end{equation}
where $e_3$ is the 3rd standard basis of $\mathbb{R}^3$. A similar linear equation is derived for the lower bound, and the depth constraints can be trivially shown to be linear.

\subsection{Quadratic Formulation for Minimum Variation Spline Generation}

We use the Minimum Variation Curve (MVC) objective, developed in \cite{moreton1992minimum}, as our regularization loss. As visualized in Fig. \ref{pipeline} (c), minimizing this loss produces smooth curves with realistic shapes. In our previous work, we demonstrated that these curves can be used to produce accurate thread reconstructions \cite{joglekar2023suture}.
    
Minimizing the MVC loss for a 3D curve is not straightforward. For an arbitrary parameter $s$, the curvature vector equation and MVC loss function are as follows \cite{o2006elementary, moreton1992minimum}:
\begin{equation} \label{E:curvature_eq}
    \kappa(s) = \frac{B''(s)}{\| B'(s) \|^2} + \frac{B'(s)}{\| B'(s) \|^4} \left(B''(s) \cdot B'(s)\right)
\end{equation}
\begin{equation} \label{E:mvc_arbitrary_s}
    \mathcal{L}(B(s)) = \int_0^1 \frac{\| \kappa'(s) \|^2}{\| B'(s) \|} ds
\end{equation}
Equation \eqref{E:mvc_arbitrary_s} is intractable to minimize, especially given its integrand includes the derivative of \eqref{E:curvature_eq}. We mitigated this in \cite{joglekar2023suture} by only minimizing this loss along the $z$-component of the curve, $B_z(s)$, but even so the result was computationally expensive. In this section, we demonstrate how we can drastically simplify $\mathcal{L}(B(s))$ to form a quadratic objective.

\subsubsection{Utilizing Constant-Speed Parameterization}

Instead of using an arbitrary parameter, we require $s$ to be a constant-speed parameter, i.e. $\| B'(s) \| = c$ where $c$ is some constant. Under these conditions, the second term of \eqref{E:curvature_eq} disappears because $B''(s)$ is perpendicular to $B'(s)$, and the remaining term simplifies to $\kappa(s) = c^{-2} B''(s)$. In addition, the denominator in the integrand of \eqref{E:mvc_arbitrary_s} becomes a constant. Combining these results and factoring out constants, we can rewrite \eqref{E:reconstr_formulation} as follows:
\begin{equation}\label{E:mvc}
\begin{aligned} [b]
    &\argmin_{B(s)} \int_{0}^{1} \| B'''(s) \|^2 ds\\
    \textrm{s.t. }&\forall o_j ~\exists s_j : B(s_j) \in \mathbf{R}(o_j)\\
     &\| B'(s) \| = c
\end{aligned}
\end{equation}

$\mathcal{L}(B(s))$ is now significantly simpler than \eqref{E:mvc_arbitrary_s}. In addition, it can be further reduced to a quadratic problem, as follows.

\subsubsection{Formulating a Quadratic Objective with Cubic B-Splines}

Cubic splines have been used to represent curves in a wide range of applications, due to their ability to intuitively and efficiently form complex shapes. In our case, they also further simplify our thread reconstruction problem.

We define $B(s)$ to be a cubic B-spline as follows \cite{de1978practical}:
\begin{equation} \label{E:spline_def}
    B(s) = \sum_{k = 1}^m \mathcal{B}_{k,d}(s) P_k
\end{equation}
where $P_1, \dots, P_m \in \mathbb{R}^3$ are the spline control points, $\mathcal{B}_{1,d}(s), \dots, \mathcal{B}_{m,d}(s)$ are spline basis functions, defined in \cite{de1978practical}, and $d = 3$ is the spline degree. The basis functions operate on the spline knots $t_1, \dots, t_{m+d+1} \in [0, 1]$. We require our knots to be clamped and uniform (i.e. all knots are evenly spaced, except the first and last knots have a multiplicity of $d+1$).
    
The derivative of a spline is also a spline, defined as follows \cite{mtuDerivativesBspline}:
\begin{equation} \label{E:spline_deriv_1}
    P'_k = \frac{d}{t_{k+d+1} - t_{k+1}} \left( P_{k+1} - P_k \right)
\end{equation}
\begin{equation} \label{E:spline_deriv_2}
    B'(s) = \sum_{k = 1}^{m-1} \mathcal{B}_{k+1,d-1}(s) P'_k
\end{equation}
Where $P'_k$ is our notation for the $k$-th control point of $B'(s)$, not the derivative of $P_k$. Note that for any differentiable spline with arbitrary degree $d$ and $m$ control points, its derivative has degree $d-1$, $m-1$ control points that are linear combinations of the original control points, and $m+d-1$ knots that are similarly clamped and uniform.

Utilizing the above spline properties, we can rewrite $\mathcal{L}(B(s))$ as a quadratic objective. By recursively applying \eqref{E:spline_deriv_1} and \eqref{E:spline_deriv_2}, we find that $B'''(s)$ is a degree-0 spline; this means that on any knot interval $[t'''_{k}, t'''_{k+1})$, $B'''(s) = P'''_k$. In addition, the multiplicities of the first and last knots become 1, so every knot interval $[t'''_{k}, t'''_{k+1})$ is the same size. As a result, we can reduce $\mathcal{L}(B(s))$ as follows:
\begin{equation} \label{E:mvs_sum_squares}
\begin{aligned}[b]
    \int_{0}^{1} \| B'''(s) \|^2 ds &= \sum_{k=1}^{m-3} \int_{t'''_k}^{t'''_{k+1}} \| P'''_k \|^2 ds\\
    & \propto \sum_{k=1}^{m-3} \| P'''_k \|^2 = \mathbf{P'''} \cdot \mathbf{P'''}
\end{aligned}
\end{equation}
where $\mathbf{P'''} = [\left(P_1'''\right)^\top \cdots \left(P_{m-3}'''\right)^\top]^\top \in \mathbb{R}^{3(m-3)}$ is the vectorization of the control points of $B'''(s)$. As shown in line \ref{alg:qpmatrix} of Algorithm \ref{alg:mvs}, since the control points of $B'''(s)$ are linear combinations of the control points of $B(s)$ and \eqref{E:mvs_sum_squares} is always non-negative we can construct a symmetric positive semi-definite matrix $\mathbf{A} \in \mathbb{R}^{3m \times 3m}$ to rewrite \eqref{E:mvs_sum_squares} as:
\begin{equation} \label{E:mvs_obj}
    \mathbf{P'''} \cdot \mathbf{P'''} = \mathbf{P}^\top \mathbf{A} \mathbf{P}
\end{equation}
where $\mathbf{P} = [P_1^\top \cdots P_m^\top]^\top \in \mathbb{R}^{3m}$ is the vectorization of the control points of $B(s)$. The exact definition of $\mathbf{A}$ is omitted for brevity, but the interested reader can derive this from \eqref{E:spline_deriv_1}.

Since \eqref{E:rel_region} is linear w.r.t. $B(s_j)$ and \eqref{E:spline_def} is linear w.r.t. the control points, we find that the reliability-region constraints are also linear w.r.t. the control points:
\begin{equation} \label{E:rel_region_spline}
    B(s_j) \in \mathbf{R}(o_j) \iff C_j \mathbf{P} + f_j \leq \mathbf{0}
\end{equation}
Again, the exact definitions of $C_j \in \mathbb{R}^{3 \times 3m}$ and $f_j \in \mathbb{R}^{3}$ are omitted for brevity, but the interested reader can derive these from \eqref{E:rel_region}, \eqref{E:spline_def}, and the example \eqref{E:rel_linear_ex}.

With \eqref{E:mvs_obj} and \eqref{E:rel_region_spline}, we can reformulate \eqref{E:mvc} as:
\begin{equation} \label{E:mvs}
\begin{aligned} [b]
    &\argmin_{\mathbf{P}} \mathbf{P}^\top \mathbf{A} \mathbf{P}\\
    \textrm{s.t. } &\forall o_j ~\exists s_j : C_j \mathbf{P} + f_j \leq \mathbf{0}\\
    &\| B'(s) \| = c
\end{aligned}
\end{equation}
As shown, we only use the spline control points as our decision variable, requiring all other spline parameters to remain constant.

\subsection{Spline Reconstruction with Iterative Quadratic Programming}

\begin{figure}[t]
    \centering
    \vspace{2mm}
    \includegraphics[width=0.95\linewidth]{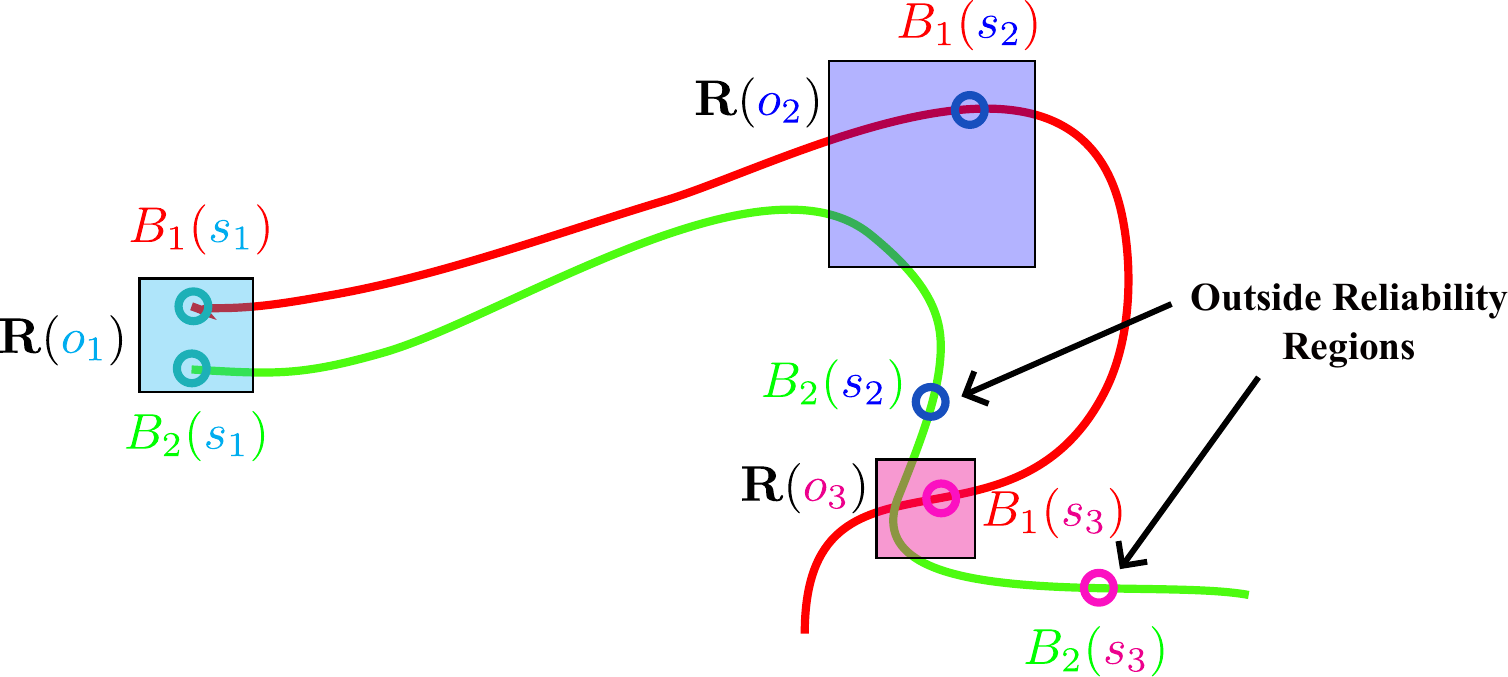}
    \caption{In this toy two-dimensional example with 3 reliability regions, we can construct two splines, $B_1(s)$ and $B_2(s)$, that pass through all regions. Let $s_1, s_2,$ and $s_3$ be a fixed set of parameters for which the reliability-region constraints are met for $B_1(s)$. If we evaluate $B_2(s)$ at these same $s_j$ values, we find that $B(s_2)$ and $B(s_3)$ lie outside the reliability regions. Hence, $s_j$ values must change when generating different splines. We adopt an iterative spline-fitting approach that allows the $s_j$ parameters to change between each iteration.}
    \label{arclen}
    \vspace{-5mm}
\end{figure}

Although \eqref{E:mvs} has a quadratic objective, it is not straightforward to solve with a generic QP solver because of its constraints. The reliability region constraints are linear for a fixed set of $s_j$ values, but, as shown in Fig. \ref{arclen}, the $s_j$ values can change based on the spline configuration. In addition, the constant-speed parameter constraint is not just nonlinear but in fact almost always impossible to satisfy \cite{farouki1991real}. To mitigate these issues, we design an algorithm that iteratively produces spline curves from quadratic subproblems, gradually approximating a valid solution to \eqref{E:mvs}. This is implemented from line \ref{alg:constructmvs_start} to \ref{alg:constructmvs_end} in Algorithm \ref{alg:mvs}.

For some iteration $i$, we define a fixed set of monotonically increasing curve parameters $s_1^{(i)} \dots s_n^{(i)} \in [0, 1]$, where $s_1^{(i)} = 0$ and $s_n^{(i)} = 1$, which we use to construct a spline curve $B^{(i)}(s)$ with control point vector $\mathbf{P}^{(i)}$. Since the $s_j^{(i)}$ parameters are fixed within the iteration, we can directly satisfy the reliability-region constraints as a set of fixed linear inequalities. We temporarily ignore the constant-speed constraint. This results in the following subproblem:
\begin{equation} \label{E:mvs_subqp}
\begin{aligned} [b]
    &\argmin_{\mathbf{P}^{(i)}} \mathbf{P}^{(i)\top} \mathbf{A} \mathbf{P}^{(i)}\\
    \textrm{s.t. } & \forall s_j^{(i)} : C_j^{(i)} \mathbf{P}^{(i)} + f_j^{(i)} \leq \mathbf{0}
\end{aligned}
\end{equation}
where each $C_j^{(i)}$ and $f_j^{(i)}$ is generated from $s_j^{(i)}$, as defined in \eqref{E:rel_region_spline}. Problem \eqref{E:mvs_subqp} is a linearly-constrained quadratic program (LCQP), a well studied optimization problem that can be efficiently solved by state-of-the-art solvers \cite{stellato2020osqp}. As implemented from line \ref{alg:qpstep_start} to \ref{alg:qpstep_end} in Algorithm \ref{alg:mvs}, upon solving \eqref{E:mvs_subqp} we obtain a candidate spline $B^{(i)}(s)$.

Next, we enforce an approximate constant-speed parameter constraint by applying the following parameter update rule between each iteration:
\begin{equation} \label{E:update_step}
    s^{(i+1)}_j = \frac{\int_0^{s^{(i)}_j} \| (B^{(i)})'(s) \| ds}{\int_0^1 \| (B^{(i)})'(s) \| ds}
\end{equation}
The derivation for \eqref{E:update_step} is detailed in the appendix, and it's implemented in line \ref{alg:updatestep} of Algorithm \ref{alg:mvs}, following the parameter initialization step implemented from line \ref{alg:initparams_start} to \ref{alg:initparams_end}. Empirically, this algorithm converges to an approximate minimum variation spline within roughly 5 iterations.

\subsection{Reliability-Driven Thread Grasping}

Given the achievement of a reliability-informed reconstruction, we can design a gripper trajectory framework that utilizes our reconstruction to enable robust grasping. Consider the case where our grasp goal point is $B(s_G)$, but our reconstruction is inaccurate at $B(s_G)$. If we directly move the gripper to $B(s_G)$, as shown in Fig. \ref{grasp}(a), then it's likely that it will miss the thread. Instead, as shown in Fig. \ref{grasp}(b), The gripper should first ``capture" the thread at a more reliable location, keeping the jaws slightly open so that it can ``slide" along $B(s)$ while keeping the thread enclosed and finally grasp the thread at $B(s_G)$. Generating such a ``capture, slide, grasp" (CSG) trajectory mainly depends on selecting the capture point $B(s_C)$, as the resulting slide waypoints can be trivially generated afterwards by discretizing $B(s)$. Note that the degenerate case where $B(s_C) = B(s_G)$ is equivalent to a direct grasp.

We utilize the reconstruction reliability regions to model the probability of a successful grasp as a function of the capture point. We pre-discretize $B(s)$ and designate all discrete parameters between $s_C$ and $s_G$ as CSG trajectory waypoints, $\mathfrak{s}_1 \dots \mathfrak{s}_w \in [0, 1]$, where $\mathfrak{s}_1 = s_C$ and $\mathfrak{s}_w = s_G$. The probability of a successful trajectory, $\mathcal{P}(\mathfrak{s}_1, \dots, \mathfrak{s}_w)$, can be written as follows:
\begin{equation} \label{E:csg_prob}
    \mathcal{P}(\mathfrak{s}_1, \dots, \mathfrak{s}_w) = \mathcal{P}(\mathfrak{s}_1) \prod_{x=2}^w \mathcal{P}(\mathfrak{s}_x | \mathfrak{s}_{x-1})
\end{equation}
where $\mathcal{P}(\mathfrak{s}_1)$ is the probability that capturing $B(\mathfrak{s}_1)$ is successful and $\mathcal{P}(\mathfrak{s}_x | \mathfrak{s}_{x-1})$ is the probability that the thread stays enclosed when the gripper slides from $\mathfrak{s}_{x-1}$ to $\mathfrak{s}_x$. Note that we assume the success of each waypoint depends at most on its previous waypoint. $\mathcal{P}(\mathfrak{s}_1)$ depends on the reconstruction confidence at $B(\mathfrak{s}_1)$, which is inversely related to the size of the nearby reliability regions. Hence, by linearly interpolating the $\mathbf{R}(o_j)$ regions near $B(\mathfrak{s}_1)$ we can estimate $\mathcal{P}(\mathfrak{s}_1)$. We estimate $\mathcal{P}(\mathfrak{s}_x | \mathfrak{s}_{x-1})$ as a constant close to 1, as the thread is unlikely to slip out of the gripper between the evenly-spaced waypoints.

Combining these results, our grasping policy selects the capture point that maximizes the probability of a successful CSG trajectory, as follows:
\begin{equation} \label{E:grasp_policy}
    \pi (B(s), \mathbf{R}(o_j)) = \argmax_{\mathfrak{s}_1 \in [0, 1]} \mathcal{P}(\mathfrak{s}_1) (\mathcal{P}(\mathfrak{s}_x | \mathfrak{s}_{x-1}))^{w-1}
\end{equation}

\section{EXPERIMENTS \& RESULTS}

\begin{table*}[t]
    \centering
    \vspace{2mm}
    \caption{Grasping Success Rate}
    \label{results}
    \begin{tabular}{c|cc|cc|cc|cc|cc|c}
        \toprule
        Test Case & \multicolumn{2}{c|}{Easy}
        & \multicolumn{2}{c|}{Medium} & \multicolumn{2}{c|}{Hard} & \multicolumn{2}{c|}{Singularity} & \multicolumn{2}{c|}{Occlusion} & \multirow{2}{*}{Total}\\
        Background & Paper & Chicken & Paper & Chicken & Paper & Chicken & Paper & Chicken & Paper & Chicken &\\
        \hline
        Direct Grasping &95.2\% & \textbf{100\%} & 90\% & \textbf{100\%} & 90\% & 71.4\% & 71.4\% & 90\% & \textbf{100\%} & \textbf{100\%} & 90.5\%\\
        \hline
        Robust Grasping &\textbf{100\%} & \textbf{100\%} & \textbf{94.4\%} & 95.2\% & \textbf{100\%} & \textbf{100\%} & \textbf{100\%} & \textbf{100\%} & \textbf{100\%} & 80\% & \textbf{97.0\%}\\
        \bottomrule
    \end{tabular}
\end{table*}

\begin{figure*}[t]
    \centering
    \includegraphics[width=0.98\linewidth]{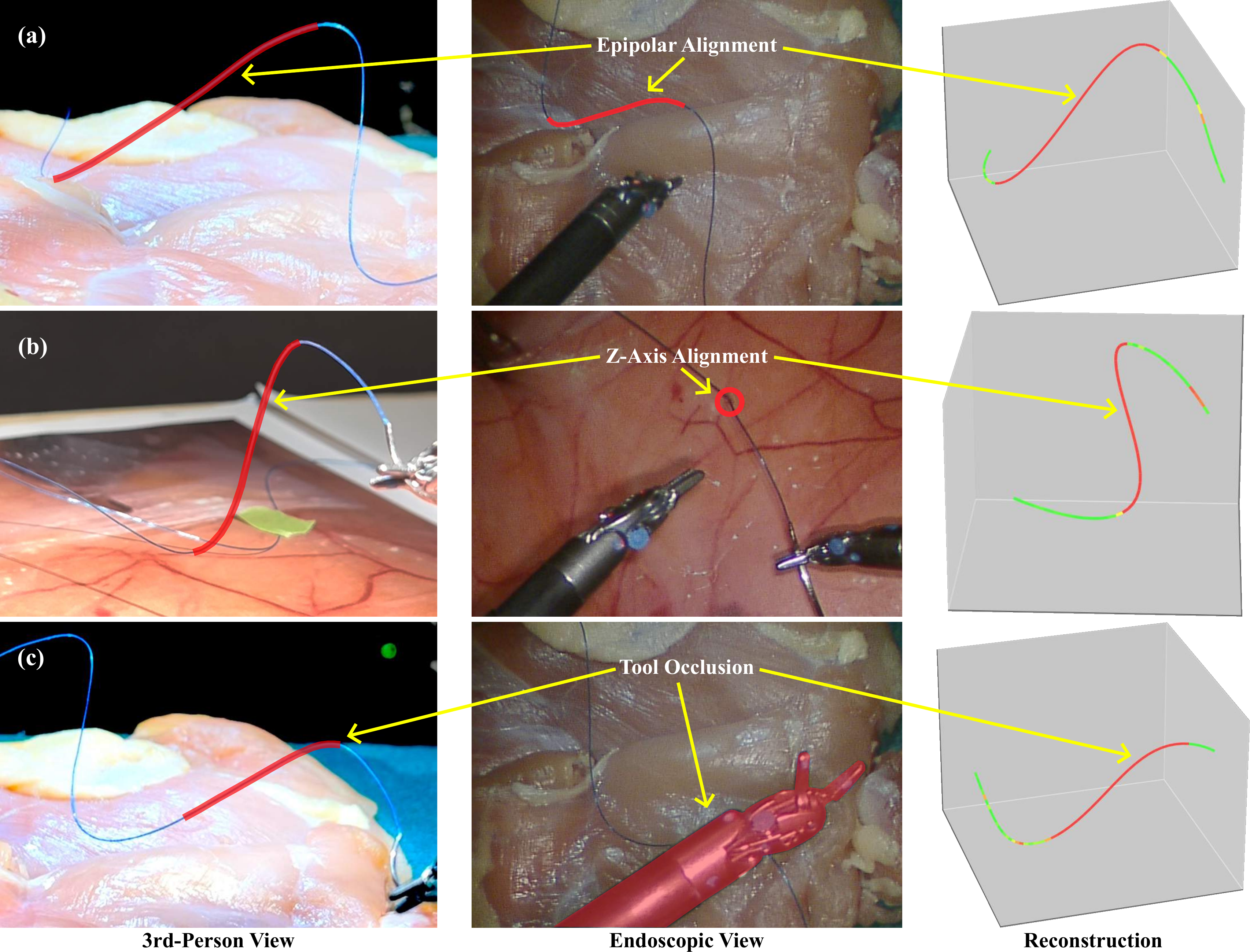}
    \caption{We crafted 10 evaluative scenarios, consisting of 5 thread configuration types with each repeated on 2 backgrounds. The 3rd-person and endoscopic views and reconstructions of 3 particular scenarios are displayed here: (a) the Medium configuration with chicken background which tests for moderate thread alignment with epipolar lines, (b) the Singularity configuration with surgical paper background which tests for thread alignment with the endoscope's $z$-axis, and (c) the Occlusion scenario with chicken background which tests for tool occlusion. Our image-to-grasp pipeline not only produces accurate reconstructions but also correctly estimates reconstruction confidence based on each perceptual degradation case.}
    \label{reconstr_exp}
\end{figure*}

\begin{figure*}[t]
    \centering
    \vspace{2mm}
    \includegraphics[width=0.98\linewidth]{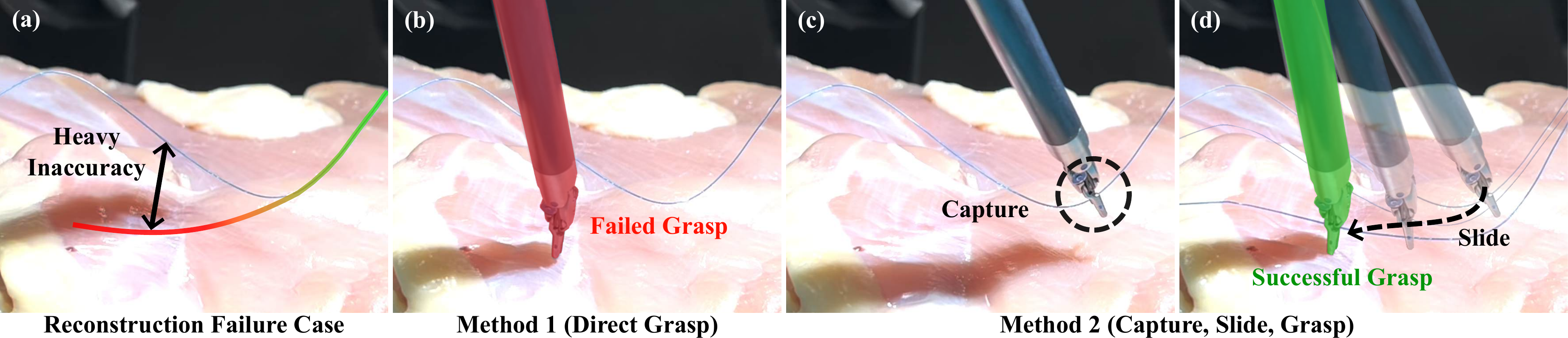}
    \caption{(a) In the experiment scenario with a Hard configuration on a chicken background, our reconstruction was inaccurate in a large portion of the thread due to heavy epipolar alignment. (b) Attempting direct grasps (Method 1) in this portion failed. However, our robust grasping policy (Method 2) was able to (c) ``capture" the thread at a reliable location and (d) execute a ``slide" trajectory to successfully grasp points in the inaccurate region, keeping the thread within the gripper jaw throughout its motion.}
    \label{grasp_exp}
    \vspace{-5mm}
\end{figure*}

We evaluate our image-to-grasp pipeline in a series of thread grasping experiments over multiple thread configurations, conducting over 400 grasping trials. We use HQ-SAM for thread segmentation from endoscopic images \cite{ke2024segment}. This model was not finetuned, demonstrating the generality of our method across various backgrounds. We use OpenCV and SciPy functions for stereo rectification and spline generation \cite{bradski2000opencv, virtanen2020scipy, dierckx1982algorithms, dierckx1995curve}. We use $m = 20$ control points for each spline reconstruction and run the reconstruction algorithm for 5 iterations, using OSQP as our quadratic programming solver and 6th order Gaussian-Legendre quadrature to approximate the integrals in \eqref{E:update_step} \cite{stellato2020osqp, abramowitz1968handbook}. We set $\mathcal{P}(\mathfrak{s}_x | \mathfrak{s}_{x-1}) = 0.99$ and estimate $\mathcal{P}(\mathfrak{s}_1)$ by linearly interpolating the $\epsilon_{j, z}$ values of each $\mathbf{R}(o_j)$ and plugging the result into a gaussian, such that the capture probability reduces as region size increases. We implement \eqref{E:grasp_policy} by sampling and evaluating 100 evenly spaced points along $B(s)$, which are reused when selecting slide trajectory waypoints. We define our gripper coordinate frame similarly to \cite{lu2021toward}, requiring waypoint poses to align the jaw's rotational axis with the curve tangent, $B'(s)$.

We conducted experiments with the da Vinci Research Kit (dVRK) surgical robotic system, collecting stereo images from the Endoscopic Camera Manipulator (ECM) and grasping with the Patient Side Manipulator (PSM) equipped with a Large Needle Driver (LND). We used the particle filter implemented in \cite{richter2021robotic} for PSM tool-tip tracking.

\subsection{Experiment Setup}

We crafted 5 thread configurations: Easy, Medium, Hard, Singularity, and Occlusion. Stereo matching is generally more difficult in regions where the thread is aligned with or frequently intersects with an epipolar line of the stereo camera, as shown in Fig. \ref{reconstr_exp}(a) \cite{joglekar2023suture}. The first three thread configurations are organized in increasing difficulty, such that the Hard configuration has significantly more epipolar alignment than the Easy configuration. The Singularity configuration includes a large section of thread that is aligned with the camera's $z$-axis, such that it is only visible in a small block of pixels in each image as shown in Fig. \ref{reconstr_exp}(b). Lastly, in the Occlusion configuration we used the PSM tool to occlude a large portion in the middle of the thread during reconstruction, shown in Fig, \ref{reconstr_exp}(c). Each configuration was replicated on 2 backgrounds, a printed surgical scene on paper and a real chicken tissue, resulting in 10 total scenarios.

For each scenario, we conducted grasping trials along the entire length of the visible thread. Each thread was discretized into approximately 20 evenly-spaced grasping points. These grasping points were used to evaluate 2 grasping strategies: direct grasping and robust CSG trajectory grasping. For each strategy, we attempted to grasp all grasping points, recording successes and failures. Over all scenarios, we perform approximately 200 grasping trials for each grasping strategy, resulting in over 400 total data points.

\subsection{Grasping Results}

Our results are detailed in Table \ref{results}, with specific reconstruction and grasping examples displayed in Fig. \ref{reconstr_exp} and Fig. \ref{grasp_exp}. Direct grasping achieves an overall 90.5\% success rate. Since direct grasping solely relies on the reconstruction, these results suggest our reconstruction by itself tends to be highly accurate. However, in a couple cases, such as the Hard-Chicken configuration, extreme reconstruction inaccuracies caused clustered grasping failures, as displayed in Fig. \ref{grasp_exp}(a, b). Grasping with CSG trajectories achieves a higher overall success rate of 97.0\%. This demonstrates that estimating and leveraging reliability information drastically improves suture thread grasping, especially in the previously discussed case where large portions of the reconstruction are inaccurate as shown in Fig. \ref{grasp_exp} (c, d). The most robust grasp failures occurred in the Occlusion-Chicken scenario when the gripper failed to securely capture the thread. Additionally, robust grasps executed in 8.0 seconds on average.


\section{CONCLUSION}

We present a reliability-driven suture thread image-to-grasp pipeline. We develop a new thread reconstruction algorithm, utilizing QP subproblems to iteratively approximate a minimum variation spline subject to reliability bounds. We then design a ``capture, slide, grasp" framework with a policy that uses reconstruction confidence to maximize grasp probability. Both components were demonstrated to have state-of-the-art accuracy through over 400 grasping trials.

Our image-to-grasp pipeline can be further improved when combined with previous works. Lu et al. introduces a ``transverse" action where the gripper attempts to grasp and move the thread, using a SIMI vector similarity metric to evaluate whether or not the thread successfully moves \cite{lu2021toward}. This enables the robot to detect and retry failed grasps until they succeed, even when reconstruction error exceeds 10 mm. This can be appended to our pipeline to further robustify grasping. Additionally, Jackson et al. and Schorp et al. both present methods for tracking suture thread over multiple frames with splines \cite{jackson2017real, schorp2023self}. Our reconstruction confidence estimation can augment their tracking pipelines with probabilistic information, enabling more sophisticated frame update steps.

The major drawback of our work is its current inability to handle self-intersecting thread configurations. However, this can be solved by relatively minor edits to our image-to-grasp pipeline, particularly the process of ordering observations, utilizing methods from other DLO reconstruction works \cite{shivakumar2023sgtm}. 

Our method opens the door for developing suture thread manipulation techniques necessary for autonomous suturing. Future work includes using our reconstruction to plan entanglement-free gripper paths during needle manipulation. In addition, our full image-to-grasp pipeline can be integrated with current suture knot-tying approaches for increased generality. Hence, our work can have significant impacts for advancing autonomous suturing procedures.

\section{APPENDIX}

By the Fundamental Theorem of Calculus and the limit definition of the derivative, we rewrite the constant-speed constraint, $\| B'(s) \| = c$, as follows:
\begin{equation} \label{E:const_speed_rewrite}
    c = \lim_{\delta \to 0} \frac{\int_s^{s+\delta} \| B'(u) \| du}{\delta}
\end{equation}
Instead of constraining the instantaneous speed for all $s$, we constrain the average speed evaluated between each consecutive $s_j$ parameter:
\begin{equation} \label{E:const_speed_approx}
    c = \frac{\int_{s_j}^{s_j+\Delta s_j} \| B'(s) \| ds}{\Delta s_j}
\end{equation}
where $\Delta s_j = s_{j+1} - s_j$.

Next, we perform 3 manipulations: we rewrite $c$ as the average velocity of the entire curve, switch $c$ and $\Delta s_j$ in \eqref{E:const_speed_approx}, and cumulatively sum each instance of \eqref{E:const_speed_approx}:
\begin{equation} \label{E:cumulative_sj}
    \sum_{l=1}^{j-1} \Delta s_l = \sum_{l=1}^{j-1} \frac{\int_{s_l}^{s_l+\Delta s_l} \| B'(s) \| ds}{\int_0^1 \| B'(s) \| ds}
\end{equation}
Given that $s_1 = 0$, we simplify \eqref{E:cumulative_sj} as follows:
\begin{equation} \label{E:direct_sj}
    s_j = \frac{\int_{0}^{s_j} \| B'(s) \| ds}{\int_0^1 \| B'(s) \| ds}
\end{equation}
This iteration-agnostic constraint is translated into the iterative update rule in Equation \eqref{E:update_step}. In addition, note that to initialize $s_1^{(1)} \dots s_n^{(1)}$ we linearly interpolate our observations and similarly constrain the average speed between them, as implemented from line \ref{alg:initparams_start} to \ref{alg:initparams_end} in Algorithm \ref{alg:mvs}.




\balance
\bibliographystyle{IEEEtran}
\bibliography{root}

\begin{thebibliography}{10}
\providecommand{\url}[1]{#1}
\csname url@rmstyle\endcsname
\providecommand{\newblock}{\relax}
\providecommand{\bibinfo}[2]{#2}
\providecommand\BIBentrySTDinterwordspacing{\spaceskip=0pt\relax}
\providecommand\BIBentryALTinterwordstretchfactor{4}
\providecommand\BIBentryALTinterwordspacing{\spaceskip=\fontdimen2\font plus
\BIBentryALTinterwordstretchfactor\fontdimen3\font minus \fontdimen4\font\relax}
\providecommand\BIBforeignlanguage[2]{{%
\expandafter\ifx\csname l@#1\endcsname\relax
\typeout{** WARNING: IEEEtran.bst: No hyphenation pattern has been}%
\typeout{** loaded for the language `#1'. Using the pattern for}%
\typeout{** the default language instead.}%
\else
\language=\csname l@#1\endcsname
\fi
#2}}

\bibitem{lin2023superpm}
S.~Lin, A.~J. Miao, A.~Alabiad, F.~Liu, K.~Wang, J.~Lu, F.~Richter, and M.~C. Yip, ``Superpm: A large deformation-robust surgical perception framework based on deep point matching learned from physical constrained simulation data,'' \emph{arXiv preprint arXiv:2309.13863}, 2023.

\bibitem{shinde2024jiggle}
N.~U. Shinde, X.~Liang, F.~Liu, Y.~Zhang, F.~Richter, S.~Herbert, and M.~C. Yip, ``Jiggle: An active sensing framework for boundary parameters estimation in deformable surgical environments,'' \emph{arXiv preprint arXiv:2405.09743}, 2024.

\bibitem{richter2021autonomous}
F.~Richter, S.~Shen, F.~Liu, J.~Huang, E.~K. Funk, R.~K. Orosco, and M.~C. Yip, ``Autonomous robotic suction to clear the surgical field for hemostasis using image-based blood flow detection,'' \emph{IEEE Robotics and Automation Letters}, vol.~6, no.~2, pp. 1383--1390, 2021.

\bibitem{miao2024hemoset}
A.~J. Miao, S.~Lin, J.~Lu, F.~Richter, B.~Ostrander, E.~K. Funk, R.~K. Orosco, and M.~C. Yip, ``Hemoset: The first blood segmentation dataset for automation of hemostasis management,'' in \emph{2024 International Symposium on Medical Robotics (ISMR)}.\hskip 1em plus 0.5em minus 0.4em\relax IEEE, 2024, pp. 1--7.

\bibitem{hubens2003performance}
G.~Hubens, H.~Coveliers, L.~Balliu, M.~Ruppert, and W.~Vaneerdeweg, ``A performance study comparing manual and robotically assisted laparoscopic surgery using the da vinci system,'' \emph{Surgical Endoscopy and other interventional techniques}, vol.~17, pp. 1595--1599, 2003.

\bibitem{ostrander2024current}
B.~T. Ostrander, D.~Massillon, L.~Meller, Z.-Y. Chiu, M.~Yip, and R.~K. Orosco, ``The current state of autonomous suturing: a systematic review,'' \emph{Surgical Endoscopy}, vol.~38, no.~5, pp. 2383--2397, 2024.

\bibitem{chiu2021bimanual}
Z.-Y. Chiu, F.~Richter, E.~K. Funk, R.~K. Orosco, and M.~C. Yip, ``Bimanual regrasping for suture needles using reinforcement learning for rapid motion planning,'' in \emph{2021 IEEE International Conference on Robotics and Automation (ICRA)}.\hskip 1em plus 0.5em minus 0.4em\relax IEEE, 2021, pp. 7737--7743.

\bibitem{chiu2023real}
Z.-Y. Chiu, F.~Richter, and M.~C. Yip, ``Real-time constrained 6d object-pose tracking of an in-hand suture needle for minimally invasive robotic surgery,'' in \emph{2023 IEEE International Conference on Robotics and Automation (ICRA)}.\hskip 1em plus 0.5em minus 0.4em\relax IEEE, 2023, pp. 4761--4767.

\bibitem{hari2024stitch}
K.~Hari, H.~Kim, W.~Panitch, K.~Srinivas, V.~Schorp, K.~Dharmarajan, S.~Ganti, T.~Sadjadpour, and K.~Goldberg, ``Stitch: Augmented dexterity for suture throws including thread coordination and handoffs,'' \emph{arXiv preprint arXiv:2404.05151}, 2024.

\bibitem{kim2024surgical}
J.~W. Kim, T.~Z. Zhao, S.~Schmidgall, A.~Deguet, M.~Kobilarov, C.~Finn, and A.~Krieger, ``Surgical robot transformer (srt): Imitation learning for surgical tasks,'' \emph{arXiv preprint arXiv:2407.12998}, 2024.

\bibitem{jiang2023markerless}
Y.~Jiang, H.~Zhou, and G.~S. Fischer, ``Markerless suture needle tracking from a robotic endoscope based on deep learning,'' in \emph{2023 International Symposium on Medical Robotics (ISMR)}.\hskip 1em plus 0.5em minus 0.4em\relax IEEE, 2023, pp. 1--7.

\bibitem{wilcox2022learning}
A.~Wilcox, J.~Kerr, B.~Thananjeyan, J.~Ichnowski, M.~Hwang, S.~Paradis, D.~Fer, and K.~Goldberg, ``Learning to localize, grasp, and hand over unmodified surgical needles,'' in \emph{2022 International Conference on Robotics and Automation (ICRA)}.\hskip 1em plus 0.5em minus 0.4em\relax IEEE, 2022, pp. 9637--9643.

\bibitem{lu2021dual}
S.~Lu, T.~Shkurti, and M.~C. {\c{C}}avușo{\u{g}}lu, ``Dual-arm needle manipulation with the da vinci{\textregistered} surgical robot under uncertainty,'' in \emph{2021 IEEE International Conference on Robotics and Automation (ICRA)}.\hskip 1em plus 0.5em minus 0.4em\relax IEEE, 2021, pp. 7744--7750.

\bibitem{chow2014novel}
D.-L. Chow, R.~C. Jackson, M.~C. {\c{C}}avu{\c{s}}o{\u{g}}lu, and W.~Newman, ``A novel vision guided knot-tying method for autonomous robotic surgery,'' in \emph{2014 IEEE international conference on automation science and engineering (CASE)}.\hskip 1em plus 0.5em minus 0.4em\relax IEEE, 2014, pp. 504--508.

\bibitem{geiger2010efficient}
A.~Geiger, M.~Roser, and R.~Urtasun, ``Efficient large-scale stereo matching,'' in \emph{Asian conference on computer vision}.\hskip 1em plus 0.5em minus 0.4em\relax Springer, 2010, pp. 25--38.

\bibitem{chang2018pyramid}
J.-R. Chang and Y.-S. Chen, ``Pyramid stereo matching network,'' in \emph{Proceedings of the IEEE conference on computer vision and pattern recognition}, 2018, pp. 5410--5418.

\bibitem{schulman2013tracking}
J.~Schulman, A.~Lee, J.~Ho, and P.~Abbeel, ``Tracking deformable objects with point clouds,'' in \emph{2013 IEEE International Conference on Robotics and Automation}.\hskip 1em plus 0.5em minus 0.4em\relax IEEE, 2013, pp. 1130--1137.

\bibitem{yang2022particle}
Y.~Yang, J.~A. Stork, and T.~Stoyanov, ``Particle filters in latent space for robust deformable linear object tracking,'' \emph{IEEE Robotics and Automation Letters}, vol.~7, no.~4, pp. 12\,577--12\,584, 2022.

\bibitem{liu2023robotic}
F.~Liu, E.~Su, J.~Lu, M.~Li, and M.~C. Yip, ``Robotic manipulation of deformable rope-like objects using differentiable compliant position-based dynamics,'' \emph{IEEE Robotics and Automation Letters}, vol.~8, no.~7, pp. 3964--3971, 2023.

\bibitem{jackson2017real}
R.~C. Jackson, R.~Yuan, D.-L. Chow, W.~S. Newman, and M.~C. {\c{C}}avu{\c{s}}o{\u{g}}lu, ``Real-time visual tracking of dynamic surgical suture threads,'' \emph{IEEE Transactions on Automation science and Engineering}, vol.~15, no.~3, pp. 1078--1090, 2017.

\bibitem{schorp2023self}
V.~Schorp, W.~Panitch, K.~Shivakumar, V.~Viswanath, J.~Kerr, Y.~Avigal, D.~M. Fer, L.~Ott, and K.~Goldberg, ``Self-supervised learning for interactive perception of surgical thread for autonomous suture tail-shortening,'' in \emph{2023 IEEE 19th International Conference on Automation Science and Engineering (CASE)}.\hskip 1em plus 0.5em minus 0.4em\relax IEEE, 2023, pp. 1--6.

\bibitem{lu2021toward}
B.~Lu, B.~Li, W.~Chen, Y.~Jin, Z.~Zhao, Q.~Dou, P.-A. Heng, and Y.~Liu, ``Toward image-guided automated suture grasping under complex environments: A learning-enabled and optimization-based holistic framework,'' \emph{IEEE Transactions on Automation Science and Engineering}, vol.~19, no.~4, pp. 3794--3808, 2021.

\bibitem{joglekar2023suture}
N.~Joglekar, F.~Liu, R.~Orosco, and M.~Yip, ``Suture thread spline reconstruction from endoscopic images for robotic surgery with reliability-driven keypoint detection,'' in \emph{2023 IEEE International Conference on Robotics and Automation (ICRA)}.\hskip 1em plus 0.5em minus 0.4em\relax IEEE, 2023, pp. 4747--4753.

\bibitem{moreton1992minimum}
H.~P. Moreton, \emph{Minimum curvature variation curves, networks, and surfaces for fair free-form shape design}.\hskip 1em plus 0.5em minus 0.4em\relax University of California, Berkeley, 1992.

\bibitem{o2006elementary}
B.~O'neill, \emph{Elementary differential geometry}.\hskip 1em plus 0.5em minus 0.4em\relax Elsevier, 2006.

\bibitem{de1978practical}
C.~De~Boor and C.~De~Boor, \emph{A practical guide to splines}.\hskip 1em plus 0.5em minus 0.4em\relax springer New York, 1978, vol.~27.

\bibitem{mtuDerivativesBspline}
``{D}erivatives of a {B}-spline {C}urve --- pages.mtu.edu,'' \url{https://pages.mtu.edu/~shene/COURSES/cs3621/NOTES/spline/B-spline/bspline-derv.html}, [Accessed 30-07-2024].

\bibitem{farouki1991real}
R.~T. Farouki and T.~Sakkalis, ``Real rational curves are not ‘unit speed’,'' \emph{Computer Aided Geometric Design}, vol.~8, no.~2, pp. 151--157, 1991.

\bibitem{stellato2020osqp}
B.~Stellato, G.~Banjac, P.~Goulart, A.~Bemporad, and S.~Boyd, ``Osqp: An operator splitting solver for quadratic programs,'' \emph{Mathematical Programming Computation}, vol.~12, no.~4, pp. 637--672, 2020.

\bibitem{ke2024segment}
L.~Ke, M.~Ye, M.~Danelljan, Y.-W. Tai, C.-K. Tang, F.~Yu, \emph{et~al.}, ``Segment anything in high quality,'' \emph{Advances in Neural Information Processing Systems}, vol.~36, 2024.

\bibitem{bradski2000opencv}
G.~Bradski, ``The opencv library.'' \emph{Dr. Dobb's Journal: Software Tools for the Professional Programmer}, vol.~25, no.~11, pp. 120--123, 2000.

\bibitem{virtanen2020scipy}
P.~Virtanen, R.~Gommers, T.~E. Oliphant, M.~Haberland, T.~Reddy, D.~Cournapeau, E.~Burovski, P.~Peterson, W.~Weckesser, J.~Bright, \emph{et~al.}, ``Scipy 1.0: fundamental algorithms for scientific computing in python,'' \emph{Nature methods}, vol.~17, no.~3, pp. 261--272, 2020.

\bibitem{dierckx1982algorithms}
P.~Dierckx, ``Algorithms for smoothing data with periodic and parametric splines,'' \emph{Computer Graphics and Image Processing}, vol.~20, no.~2, pp. 171--184, 1982.

\bibitem{dierckx1995curve}
------, \emph{Curve and surface fitting with splines}.\hskip 1em plus 0.5em minus 0.4em\relax Oxford University Press, 1995.

\bibitem{abramowitz1968handbook}
M.~Abramowitz and I.~A. Stegun, \emph{Handbook of mathematical functions with formulas, graphs, and mathematical tables}.\hskip 1em plus 0.5em minus 0.4em\relax US Government printing office, 1968, vol.~55.

\bibitem{richter2021robotic}
F.~Richter, J.~Lu, R.~K. Orosco, and M.~C. Yip, ``Robotic tool tracking under partially visible kinematic chain: A unified approach,'' \emph{IEEE Transactions on Robotics}, vol.~38, no.~3, pp. 1653--1670, 2021.

\bibitem{shivakumar2023sgtm}
K.~Shivakumar, V.~Viswanath, A.~Gu, Y.~Avigal, J.~Kerr, J.~Ichnowski, R.~Cheng, T.~Kollar, and K.~Goldberg, ``Sgtm 2.0: Autonomously untangling long cables using interactive perception,'' in \emph{2023 IEEE International Conference on Robotics and Automation (ICRA)}.\hskip 1em plus 0.5em minus 0.4em\relax IEEE, 2023, pp. 5837--5843.

\end{thebibliography}

\end{document}